\documentclass[10pt,conference]{IEEEtran}
\usepackage{cite}
\usepackage{amsmath,amssymb,amsfonts}
\usepackage{graphicx}
\usepackage{textcomp}
\usepackage{xcolor}
\usepackage{tikz}
\usepackage{booktabs}
\usepackage{multirow}
\usepackage{xcolor} 
\usepackage{algorithm}      
\usepackage{algpseudocode}  
\usepackage{subcaption} 
\usetikzlibrary{arrows.meta, positioning, shapes.geometric}
\def\BibTeX{{\rm B\kern-.05em{\sc i\kern-.025em b}\kern-.08em
    T\kern-.1667em\lower.7ex\hbox{E}\kern-.125emX}}
\begin{document}

\title{LogiPart: Local Large Language Models for Data Exploration at Scale with Logical Partitioning}

\author{\IEEEauthorblockN{Tiago Fernandes Tavares}
\IEEEauthorblockA{
\textit{Insper}\\
São Paulo, Brazil \\
tiagoft1@insper.edu.br}}

\maketitle

\begin{abstract}
The discovery of deep, steerable taxonomies in large text corpora is currently restricted by a trade-off between the surface-level efficiency of topic models and the prohibitive, non-scalable assignment costs of LLM-integrated frameworks. We introduce \textbf{LogiPart}, a scalable, hypothesis-first framework for building interpretable hierarchical partitions that decouples hierarchy growth from expensive full-corpus LLM conditioning. LogiPart utilizes locally hosted LLMs on compact, embedding-aware samples to generate concise natural-language taxonomic predicates. These predicates are then evaluated efficiently across the entire corpus using zero-shot Natural Language Inference (NLI) combined with fast graph-based label propagation, achieving constant $O(1)$ generative token complexity per node relative to corpus size. We evaluate LogiPart across four diverse text corpora (totaling $\approx$140,000 documents). Using structured manifolds for \textbf{calibration}, we identify an empirical reasoning threshold at the 14B-parameter scale required for stable semantic grounding. On complex, high-entropy corpora (Wikipedia, US Bills), where traditional thematic metrics reveal an ``alignment gap,'' inverse logic validation confirms the stability of the induced logic, with individual taxonomic bisections maintaining an average per-node routing accuracy of up to 96\%. A qualitative audit by an independent LLM-as-a-judge confirms the discovery of meaningful functional axes, such as policy intent, that thematic ground-truth labels fail to capture. LogiPart enables frontier-level exploratory analysis on consumer-grade hardware, making hypothesis-driven taxonomic discovery feasible under realistic computational and governance constraints.

\end{abstract}

\begin{IEEEkeywords}
Large Language Models (LLMs), Hierarchical Text Discovery, Natural Language Inference, Label Propagation
\end{IEEEkeywords}

\section{Introduction}
Exploration of large text corpora is a fundamental challenge in artificial intelligence, yet state-of-the-art methods typically focus on thematic recovery (identifying what a document is about) rather than taxonomic discovery (surfacing the latent logical structures that define why documents differ). To a researcher, identifying why two policy documents differ in their economic impact (e.g., \textit{public interest} vs. \textit{private interest}) is often more valuable than simply labeling them both as \textit{Government}. This motivates an exploratory objective centered on functional differentiation rather than thematic label recovery.

The focus on surface-level subject matter often forces a critical trade-off between semantic fidelity and computational scalability. For instance, topic models like BERTopic~\cite{grootendorst2022bertopic} achieve efficiency but rely on keyword lists that become semantically ambiguous as topic granularity increases~\cite{Chang2009}. Modern LLM-integrated approaches, such as TopicGPT~\cite{topicgpt2024} and LiSa~\cite{liu2025lisa}, generate human-readable descriptions but depend on per-document LLM conditioning (that is, they require $O(N)$ generative calls). This reliance on per-document generative assignment imposes a computational burden that scales linearly with corpus size, rendering the discovery of deep, functional logic, such as legislative intent or economic nature, prohibitively expensive even for moderately large datasets.

In this work, we introduce LogiPart, a scalable, hypothesis-first framework for building interpretable hierarchical text partitions that ensures growing the hierarchy remains efficient even as the corpus size increases. We cast tree construction as iterative hypothesis verification, combining embedding-aware sampling, LLM-driven splitting hypotheses, and zero-shot NLI with label propagation. This design reduces the token complexity of the LLM step to $O(1)$ per node (depending only on a fixed sample size), shifting the bulk of computation to fast discriminative NLI models and high-performance label propagation. Each node produces a single, clear division, which can be manually steered by constraining or editing the generated hypothesis. Unlike post-hoc cluster labeling, these hypotheses function as executable semantic rules at inference time. To our knowledge, ours is among the first hierarchical clustering framework that combines LLM-generated hypotheses with NLI-based label propagation to avoid per-document generative assignment. Source code is available at [will be provided after anonymous peer review].

LogiPart’s design enables a transition from proprietary APIs to locally hosted, open-weight LLMs. While O(N) frameworks are theoretically portable, their linear generative demand on every document renders on-premise execution computationally prohibitive at scale. By confining generative calls to fixed-size samples $O(1)$ per node, \textsc{LogiPart} ensures exploratory analysis remains feasible on consumer-grade hardware, aligning with critical needs for cost predictability~\cite{pan2025costbenefitanalysisonpremiselarge}, data governance~\cite{McIntosh2025}, and reproducibility~\cite{manchanda2025opensourceadvantagelarge}.

\section{Related Work}
\label{sec:related}

\textbf{LLM-Augmented Clustering:} LLMs have shown remarkable capability in summarizing and organizing text. Approaches like RAPTOR~\cite{sarthi2024raptor} and work by Qiao et al.~\cite{Qiao2025} utilize recursive bottom-up summarization to build tree structures. Similarly, LLMs have been employed to refine pre-defined clusters~\cite{pattnaik-etal-2024-improving,lin2025bagoftexts}, optimize decision trees for tabular data~\cite{liu2025llego,knauer2024}, perform clustering from expert instructions~\cite{wang-etal-2023-goal,viswanathan-etal-2024-large}, or label geometrically-induced taxonomies~\cite{petnehazi2025herculeshierarchicalembeddingbasedrecursive}. TopicGPT~\cite{topicgpt2024} and LiSa~\cite{liu2025lisa} diverge from these geometry-first methods by introducing hypothesis-driven components, where LLMs generate taxonomic criteria before document assignment. However, these frameworks still require conditioning models on every document (or large, repeated samples) to perform the assignment. This reliance on per-document generative calls creates an assignment bottleneck where token consumption scales linearly with corpus size and hierarchy depth, rendering deep exploratory analysis of massive datasets economically and practically prohibitive.

\textbf{Thematic Recovery vs. Taxonomic Discovery:} Large-scale corpora typically possess multiple orthogonal semantic axes; a system may successfully discover a functional partition (e.g., policy intent or economic nature) that is mathematically independent of, and non-redundant with, thematic ground-truth labels~\cite{Gondek2006}. This creates a distinction between \textit{thematic recovery} (the automated replication of pre-defined administrative categories) and \textit{taxonomic discovery} (the surfacing of functional dimensions, or latent \textit{semantic axes}~\cite{lucy-etal-2022-discovering}). While traditional evaluations prioritize recovery performance, landmark research suggests that statistical fit often correlates negatively with human-centric interpretability~\cite{Chang2009}. Following Marchionini's framework for exploratory search~\cite{marchionini2006}, we treat the hierarchy not as a static recovery-based classifier but as a steerable engine for open-ended sensemaking, capturing cross-cutting logic that traditional thematic models consistently overlook. 

\textbf{Zero-Shot Classification:} Our use of Natural Language Inference (NLI) for document assignment draws upon work in zero-shot text classification~\cite{Laurer2023, Laurer2023b} and hallucination detection~\cite{manakul-etal-2023-selfcheckgpt}. Unlike standard clustering, which relies solely on vector space density, NLI allows documents to be evaluated against explicit semantic hypotheses. Importantly, LLMs can be used for such~\cite{tabatabaei2024largelanguagemodelsserve}, but that implies a significant cost per document. 

\textbf{Label Propagation:} Although NLI is significantly faster and cheaper than LLMs, its use in large datasets can still become prohibitive. We further improve efficiency by applying NLI to a sample of the dataset and then obtaining the remaining answers via label propagation~\cite{zhou2003learning}, which can be efficiently implemented on GPUs~\cite{Iscen_2019_CVPR}.

\textbf{Sampling and Hierarchy Construction:} The fidelity of both LLM-driven hypothesis generation and label propagation relies on selecting a representative support set. While random sampling is unbiased, K-Means-based sampling offers superior feature space coverage, approximating geometric coresets~\cite{axiotis2024}. To further mitigate redundancy in high-density regions, we utilize the Vote-K algorithm~\cite{su2023selective}, which enforces spatial diversity by penalizing candidates in close graph neighborhoods. This geometric partitioning approach aligns with the principles of Bisecting K-Means~\cite{steinbach2000}, which establishes recursive binary splitting of the embedding space as a standard paradigm for hierarchical clustering~\cite{Wang2020}.

\section{Proposed Method}
\label{sec:method}
The proposed method for tree building follows a recursive structure as shown in Algorithm~\ref{alg:code}. The recursive function begins by selecting samples from the dataset following the procedures described in Section~\ref{sec:sampling}. After that, an LLM generates a hypothesis candidate, as discussed in Section~\ref{sec:hypothesis_generation}. Then, the hypothesis is tested over the whole dataset using NLI and label propagation, as described in Section~\ref{sec:inference}. Last, the function enters recursive calls and abides by the stopping criteria discussed in Section~\ref{sec:recursion}.

\begin{algorithm}
\caption{Recursive function to build trees. It receives a list of texts $x$, a blocklist $b$, two sampling functions, LLM-S and NLI-S, and two models, LLM and NLI. The thresholds and other parameters were omitted for simplicity.}
\begin{algorithmic}[1]   
\footnotesize
    \Function{Build}{$x$, $b$, LLM-S, NLI-S, LLM, NLI}
        \State Initialize $N$ as a new tree node.

        \State $e \gets $ Embeddings($x$)
        \State $x_{LLM} \gets$ LLM-S($x,e$),  $x_{NLI} \gets$ NLI-S($x,e$)  \Comment {\textbf{Sampling dataset}}
        
        \State running $\gets$ True, attempts $\gets$ 0
        \While {running} \Comment{\textbf{Attempt hypothesis generation until valid split or max attempts}}
            \State attempts $\gets$ attempts + 1
            \If {attempts $>$ max-attempts}
            \State \Return $N$
            \EndIf
        
            \State $h \gets$ LLM($x_{LLM}$, $b$)             \Comment {\textbf{Hypothesis generation}}

            \State $b$.append($h$)

            \State $L^\prime \gets NLI(x_{NLI}, h)$ \Comment {\textbf{Inference}}
            \State $L \gets $ LabelPropagation($L^\prime, e$)

            \If {$L$ is a valid split}
            \State running $\gets$ False
            \EndIf
        \EndWhile
        \State $x_0 \gets x[L==0]$
        \State $x_1 \gets x[L==1]$

        \State $N.$left $\gets$ \Call{Build}{$x_0$, $b$, LLM-S, NLI-S, LLM, NLI}
        \State $N.$right $\gets$ \Call{Build}{$x_1$, $b$, LLM-S, NLI-S, LLM, NLI}
        \State \Return $N$
    
    \EndFunction
\end{algorithmic}
\label{alg:code}
\end{algorithm}

\subsection{Sampling}
\label{sec:sampling}

\subsubsection{Random sampling}
The most direct way to find a sample is through random sampling. This method is fast, but it can induce bias if the sample is small.

\subsubsection{K-Means}
The representativeness of the sample can be induced by selecting $k$ medoids using the K-Means algorithm. This can be implemented using FAISS~\cite{johnson2019billion}, but it requires a prior step of calculating embeddings for each text in the collection. For this, we used Sentence Bert~\cite{reimers-gurevych-2019-sentence} and a pre-trained model\footnote{We used \textsc{sentence-transformers/paraphrase-albert-small-v2} from Hugging Face}. Where applicable, we divided the text into 150-word chunks with a 50-word overlap and used the mean embedding as the document embedding.

\subsubsection{Vote-K}
One problem with the K-Means algorithm is that it can yield more means in more dense regions of the embedding space. This can be mitigated with the Vote-K algorithm~\cite {su2023selective}, which prevents the selection of medoids that are close to each other. We reimplemented the code provided by the authors using the FAISS library.

\subsubsection{Bisecting K-Means}
The sampling strategies discussed so far do not leverage the spatial geometry induced by the text embeddings. For this, we first execute a K-Means clustering with $K=2$ to bisect the space into two regions. Then we sample within each region using K-Means. This procedure is similar to that used in BERTopic~\cite{grootendorst2022bertopic}, but with a top-down clustering schema rather than a bottom-up one.

\subsection{Hypothesis generation}
\label{sec:hypothesis_generation}
Samples are yielded to an LLM that is responsible for generating an evaluation hypothesis. The prompt (Figure~\ref{fig:prompt}) uses role-prompting and specific guardrails to prevent the LLM from focusing on surface-level keywords or entities. Our system also implements a ``blocklist'', which consists of previously generated hypotheses that were either used in higher levels of the tree (to avoid redundancy) or rejected (to avoid uninformative paths).

\begin{figure}[h]
\centering
\fcolorbox{blue}{gray!10}{%
  \parbox{0.95\linewidth}{%
\footnotesize
    \textbf{System prompt:}
    
    You are a senior taxonomist.
    
    You will be given a collection of texts. Generate a concise taxonomic hypothesis
    that is true for half of the texts, and is false for the other half. The hypothesis
    must be a single assertion that can be evaluated as true or false and refers to a single text.
    
    Observe the following rules:
    
    The hypothesis must refer to a conceptual duality within the texts. 
    
    The hypothesis will be evaluated later by another agent so it must be phrased without ambiguity.
    
    The hypothesis must account solely to the content of the texts, ignoring its positioning within the collection.
    
    The hypothesis starts with ``the text'' followed by ``denotes'', ``means'', ``focuses on'', ``communicates that'' or similar verbs.
    
    The hypothesis cannot refer to a specific word, punctuation or other grammar element being used.
    
    The hypothesis must mention a valid type of content of subject.
    
    Ignore entities (companies, people, cities).
    
    Focus on the nature of the texts, and on the human intent behind them.
    
    Use higher-level abstraction instead of using topic words that exist on the texts.
    
    
    \textbf{User prompt:}
    
    Texts: [collection]

    Avoid the following topics: [blocklist]
  }%
}

\caption{Prompt template for hypothesis generation}
\label{fig:prompt}
\end{figure}

When the bisection sampling strategy is used, the prompt is changed so that the instructions read: ``You will be given two collections of texts. Generate a concise taxonomic hypothesis that is true for collection A, and is false for collection B. The hypothesis must be a single assertion that can be evaluated as true or false and refers to a single text.''. Also, the user prompt presents Collection A and Collection B instead of ``Texts''.

To ensure that all texts fit the context window, we used collections of at most 14 texts (7 per collection in the case of bisection), which were individually cropped to the first 350 words. This method assumes that the beginning of the text carries (or at least summarizes) most of its meaning, which is true for short-form, non-fiction writing such as news, Internet messages, or encyclopedia articles, but can fail with fiction or larger manuscripts such as reports or books.

\subsection{Natural Language Inference and Label Propagation}
\label{sec:inference}
The hypothesis is then evaluated over all texts in the collection using a pre-trained NLI system\footnote{We used the pretrained \textsc{MoritzLaurer/mDeBERTa-v3-base-xnli-multilingual-nli-2mil7} from HuggingFace}. The system receives as input a premise and a hypothesis, and classifies the pair as ``entailment'', ``neutral'', or ``contradiction''. We use the text itself as the premise, and the LLM-generated hypothesis as the hypothesis. We calculate the probability of entailment $P(e)$ using the entailment and contradiction logits~\cite{manakul-etal-2023-selfcheckgpt}. If necessary, we divide the texts into 150-word chunks with a 50-word overlap. Then, we use max pooling to select the maximum $P(e)$ across all chunks. If $P(e)>0.5$, the text is labeled as ``entailment'', and is labeled as ``contradiction'' otherwise.

\textbf{Label Propagation:} Although NLI is evidently faster than LLMs for finding entailments, it can still be time-consuming in large datasets. For this reason, we apply NLI to a sample of the dataset and use label propagation~\cite{Iscen_2019_CVPR} to assign labels to the remaining texts, which is orders of magnitude faster. We found that labeling 10\% of the texts in the collection is sufficient and avoids polarization across all our test datasets.

When using the bisection method, the labels provided by NLI and label propagation can override those from K-Means clustering or be used solely to evaluate stopping criteria. If the NLI labels do not override K-Means clustering, then the method approaches a hierarchical version of BERTopic, and all splits are solely based on a geometric criterion.

\subsection{Recursion and stopping criteria}
\label{sec:recursion}
After all texts are labeled, we calculate the ratio between the ``entailment'' and the ``contradiction'' sets. If it is within an allowed range (we used $[0.1,0.9]$), then the split is confirmed. In this case, the texts labeled as ``entailment'' are assigned to the left child node, the texts labeled as ``contradiction'' are assigned to the right child node, and the procedure recurses on each child, adding the generated hypothesis to the blocklist. A split ratio outside the allowed range means that either the generated hypothesis is inherently unsuitable or that it is phrased in a way that the NLI model is unable to process correctly. Hence, the hypothesis is added to the blocklist, and the procedure continues. If the system fails to generate a suitable hypothesis in more than a maximum number of attempts (we used 10), then it is considered unable to differentiate along the given data, and the recursion returns a leaf node. Also, the recursion stops if a node has less than a pre-defined number of texts (we used 200), or if the tree reaches a maximum height (we used 6).

\section{Experiments}
\label{sec:experiments}

Our evaluation follows a three-stage progression. First, we use traditional thematic metrics on simple datasets to calibrate the model's reasoning threshold. Second, we evaluate the generalization of the induced logic to held-out data. Finally, we address the 'alignment gap' in complex domains by using Inverse Logic Validation and LLM-as-a-judge to prove that the discovered axes—while orthogonal to thematic labels—represent stable and meaningful semantic structures. We use four diverse text corpora: \textbf{AG-News} \cite{zhang2015}, representing a structured four-domain manifold; \textbf{20 Newsgroups} \cite{Lang95}, a topological stress test with high inter-class overlap; and a subset of \textbf{Wikipedia} and \textbf{US Bills} \cite{topicgpt2024}, used to evaluate the system's capacity for functional discovery in high-entropy and policy-dense environments. The entire \textsc{LogiPart} pipeline (including the local LLM, document sampling, NLI, and label propagation) was conducted locally on a single NVIDIA 4090 RTX gaming GPU with 24GB VRAM (4-bit quantization using Ollama), while the LLM-as-a-judge experiment used a cloud-based LLM leveraging its superior reasoning capabilities. In all \textsc{LogiPart} experiments, we used a maximum height of 6 to balance speed and accuracy.

\subsection{Reasoning Threshold and Calibration}
\label{sec:tree_evaluation}
Our initial ablation study aims to identify the \textit{reasoning threshold} required for stable taxonomic discovery. To achieve this, we first evaluate \textsc{LogiPart} on datasets characterized by high-separability manifolds, such as AG-News, where latent logical structures and thematic ground-truth are known to coincide. In these environments, thematic alignment serves as a necessary diagnostic to verify that the LLM can anchor its hypotheses in the document text rather than hallucinate. While thematic recovery is an insufficient measure of exploratory utility in high-entropy domains, performance on these structured manifolds serves as a calibration signal for a model’s capacity for instruction-following and semantic resolution. We therefore treat these metrics not as a final target for optimization, but as a calibration tool to filter out architectures that exhibit stochastic behavior or a failure to adhere to the safeguards and blocklist constraints.

We evaluated a spectrum of local LLMs and sampling strategies using two sets of metrics: (i) \textbf{Normalized Mutual Information (NMI)} and \textbf{Adjusted Rand Index (ARI)} to assess structural alignment; and (ii) post-hoc majority voting to compute \textbf{Accuracy (ACC)} and macro-\textbf{F1-Score} as proxies for leaf-level purity. These metrics are utilized here as calibration tools to filter out configurations that exhibit stochastic behavior or failure to follow negative constraints. For comparison, we benchmark against supervised decision tree oracles. We used two oracles: CART trained on SBERT embeddings and BoW trained on bag-of-words representations. They are limited to a maximum tree height of 6, which greatly decreases their accuracy, but allows comparison with the LogiPart-generated trees.

Results for the AG-News discovery set (Table~\ref{tab:results_agnews}) reveal a clear reasoning floor at the 14B parameter scale. Models below this threshold (3B/8B) exhibit near-zero NMI and Accuracy scores indistinguishable from random chance, indicating a failure to ground logical predicates in the document text. Conversely, \textsc{qwen3:14b} and \textsc{gpt-oss:20b} show stable, non-random alignment, with the bisection-only baseline reaching an F1 of 0.80. This confirms that the embedding geometry and the model's logical intuition tend to align in this dataset.

\begin{table}[h!]
\caption{Tree evaluations on AG-News discovery set using different models and strategies for sampling. $H$ represents the achieved tree height.}
\begin{tabular}{lllrrrrr}
\toprule
 Model & LLM-S & NLI-S & H & NMI & ARI & ACC & F1 \\
\midrule
 llama3.2:3b & kmeans & kmeans & 5 & 0.00 & 0.00 & 0.25 & 0.17 \\
 llama3.2:3b & kmeans & random & 6 & 0.00 & 0.00 & 0.26 & 0.11 \\
 llama3.2:3b & random & kmeans & 0 & 0.00 & 0.00 & 0.25 & 0.10 \\
 llama3.2:3b & random & random & 3 & 0.00 & 0.00 & 0.25 & 0.10 \\
 llama3.2:3b & votek & kmeans & 6 & 0.00 & 0.00 & 0.26 & 0.16 \\
 llama3.2:3b & votek & random & 0 & 0.00 & 0.00 & 0.25 & 0.10 \\ \hline
  llama3.1:8b & kmeans & kmeans & 6 & 0.05 & 0.01 & 0.35 & 0.33 \\
 llama3.1:8b & kmeans & random & 6 & 0.01 & 0.00 & 0.30 & 0.27 \\
 llama3.1:8b & random & kmeans & 6 & 0.02 & 0.00 & 0.34 & 0.33 \\
 llama3.1:8b & random & random & 6 & 0.01 & 0.00 & 0.30 & 0.28 \\
 llama3.1:8b & votek & kmeans & 6 & 0.07 & 0.01 & 0.42 & 0.42 \\
 llama3.1:8b & votek & random & 6 & 0.02 & 0.00 & 0.30 & 0.28 \\ \hline
 qwen3:8b & kmeans & kmeans & 6 & 0.04 & 0.01 & 0.39 & 0.38 \\
 qwen3:8b & kmeans & random & 6 & 0.01 & 0.00 & 0.31 & 0.29 \\
qwen3:8b & random & kmeans & 6 & 0.01 & 0.00 & 0.33 & 0.32 \\
 qwen3:8b & random & random & 6 & 0.01 & 0.00 & 0.32 & 0.31 \\
 qwen3:8b & votek & kmeans & 6 & 0.02 & 0.00 & 0.35 & 0.33 \\
 qwen3:8b & votek & random & 6 & 0.01 & 0.00 & 0.29 & 0.25 \\ \hline 
  qwen3:14b & kmeans & kmeans & 6 & 0.07 & 0.01 & 0.43 & 0.43 \\
 qwen3:14b & kmeans & random & 6 & 0.07 & 0.01 & 0.41 & 0.39 \\
 qwen3:14b & random & kmeans & 6 & 0.07 & 0.01 & 0.43 & 0.41 \\
 qwen3:14b & random & random & 6 & 0.03 & 0.01 & 0.34 & 0.34 \\
 qwen3:14b & votek & kmeans & 6 & 0.11 & 0.03 & 0.50 & \textbf{0.50} \\
 qwen3:14b & votek & random & 6 & 0.02 & 0.00 & 0.33 & 0.31 \\ \hline
 gpt-oss:20b & kmeans & kmeans & 6 & 0.08 & 0.02 & 0.44 & \textbf{0.44} \\
gpt-oss:20b & kmeans & random & 6 & 0.05 & 0.01 & 0.38 & 0.35 \\
 gpt-oss:20b & random & kmeans & 6 & 0.07 & 0.02 & 0.42 & 0.42 \\
 gpt-oss:20b & random & random & 6 & 0.02 & 0.00 & 0.32 & 0.31 \\
 gpt-oss:20b & votek & kmeans & 6 & 0.08 & 0.01 & 0.42 & 0.41 \\
 gpt-oss:20b & votek & random & 6 & 0.03 & 0.00 & 0.34 & 0.34 \\ \hline

 \multicolumn{3}{l}{Oracle (CART)}  & 6 & 0.13 & 0.14 & 0.53 & 0.52 \\
 \multicolumn{3}{l}{Oracle (Bow)} & 6 & 0.14 & 0.02 & 0.39 & 0.35 \\ 
 \hline 
 gpt-oss:20b & \multicolumn{2}{l}{bisection, override} & 6 & 0.07 & 0.02 & 0.45 & 0.45 \\
 gpt-oss:20b & \multicolumn{2}{l}{bisection, no override} & 6 & 0.28 & 0.06 & 0.80 & 0.80 \\ 
 qwen3:14b &\multicolumn{2}{l}{bisection, override} & 6 & 0.10 & 0.02 & 0.48 & 0.47 \\
 qwen3:14b &\multicolumn{2}{l}{bisection, no override} & 6 & 0.28 & 0.06 & 0.80 & 0.80 \\

\bottomrule
\end{tabular}

\label{tab:results_agnews}
\end{table}

To challenge this result, we extended the ablation to the 20 Newsgroups dataset (Table~\ref{tab:results_20newsgroups}), characterized by higher inter-class geometry overlap. While structural alignment decreases as expected, the 20B model maintains superior consistency over its 14B counterpart.

\begin{table}[h!]
\caption{Tree evaluations on the 20 Newsgroups discovery set using different models and strategies for sampling. $H$ represents the achieved tree height.}
\begin{tabular}{lllrrrrr}
\toprule
 Model & LLM-S & NLI-S & H & NMI & ARI & ACC & F1 \\
\midrule
gpt-oss:20b & kmeans & kmeans & 6 & 0.08 & 0.01 & 0.13 & 0.10 \\
gpt-oss:20b & votek & kmeans & 6 & 0.08 & 0.01 & 0.13 & 0.10 \\
qwen3:14b & kmeans & kmeans & 6 & 0.06 & 0.01 & 0.11 & 0.09 \\
qwen3:14b & votek & kmeans & 6 & 0.05 & 0.01 & 0.11 & 0.08 \\ \hline 
 \multicolumn{3}{l}{Oracle (CART)}  & 6 & 0.21 & 0.10 & 0.26 & 0.20 \\
 \multicolumn{3}{l}{Oracle (Bow)}  & 6 & 0.16 & 0.01 & 0.17 & 0.14 \\ \hline
 gpt-oss:20b &  \multicolumn{2}{l}{bisection, override} & 6 & 0.06 & 0.01 & 0.11 & 0.10 \\
 gpt-oss:20b & \multicolumn{2}{l}{bisection, no override} & 6 & 0.40 & 0.16 & 0.48 & 0.46 \\
 qwen3:14b &  \multicolumn{2}{l}{bisection, override}  & 6 & 0.05 & 0.01 & 0.10 & 0.08 \\
 qwen3:14b & \multicolumn{2}{l}{bisection, no override} & 6 & 0.40 & 0.16 & 0.48 & 0.46 \\
\bottomrule
\end{tabular}

\label{tab:results_20newsgroups}
\end{table}

Importantly, we do not use these results to select a ``best clusterer,'' but rather to identify the \textit{minimal capable architecture} for subsequent discovery tasks. Because the \textsc{gpt-oss:20b} model demonstrated the most robust semantic grounding and highest instruction-following stability, it was selected as the engine for our primary investigations into taxonomic discovery on the Wikipedia and US Bills datasets.

\subsection{Inference Evaluation}
To assess the interpretability and discriminative power of the induced hierarchies, we treat the trees as hierarchical classifiers on held-out test sets. Each test document is routed top-down by evaluating it against the natural-language hypotheses at each node. Upon reaching a leaf, it is assigned the majority ground-truth label from the discovery-set documents that terminated at that leaf. We first evaluate the system on \textbf{AG-News} and \textbf{20 Newsgroups}, representing relatively well-separated thematic manifolds. As shown in Table~\ref{tab:results_inference}, the induced logical rules remain valid when applied to unseen documents, demonstrating \textsc{LogiPart}'s stable semantic generalization. 

The sharp decline in F1-score for the non-override bisection baseline (dropping from 0.80 in discovery to 0.40 in inference) demonstrates that post-hoc labeling of geometric clusters is semantically brittle. However, the high F1 for the KNN oracle and the high NMI/ARI for Hercules~\cite{petnehazi2025herculeshierarchicalembeddingbasedrecursive} indicate that the ground-truth labels in these datasets are well aligned with their geometry.
In contrast, the NLI-override ensures that every document is routed according to explicit logical predicates, creating a robust logical address that generalizes to unseen documents.

\subsubsection{Scaling to Complex Functional Domains}
To evaluate \textsc{LogiPart} in environments with higher semantic entropy and complex legislative structures, we extend our evaluation to the \textbf{Wikipedia} and \textbf{US Bills} datasets. As shown in Table~\ref{tab:results_wiki}, these benchmarks reveal a significant ``alignment gap'' between our discovered taxonomies and thematic ground-truth labels compared to state-of-the-art (SOTA) frameworks like TopicGPT~\cite{topicgpt2024} and LiSa~\cite{liu2025lisa}.

\begin{table}[t!]
\centering
\caption{Evaluation on held-out sets.}
\begin{tabular} {lllrrrr}
\toprule
Dataset & Model (or LLM-S) & NMI & ARI & ACC & F1 \\
\midrule
AG-News & kmeans & 0.25 & 0.24 & 0.59 & 0.57 \\
AG-News & votek & 0.33 & 0.31 & 0.64 & \textbf{0.64} \\
AG-News & bisection, override & 0.23 & 0.25 & 0.60 & 0.60 \\
AG-News & bisection, no override & 0.15 & 0.12 & 0.45 & 0.40 \\ 
AG-News & Oracle (KNN) & & & & 0.87 \\ 
AG-News & Oracle (CART) & & & & 0.54 \\ \midrule
20 Newsgroups & kmeans & 0.07 & 0.03 & 0.28 & 0.19 \\
20 Newsgroups & votek & 0.03 & 0.03 & 0.33 & \textbf{0.30} \\
20 Newsgroups & bisection, override & 0.07 & 0.04 & 0.20 & 0.15 \\
20 Newsgroups & bisection, no override & 0.03 & 0.02 & 0.26 & 0.24 \\ 
20 Newsgroups & Oracle (KNN) & & & & 0.55 \\
20 Newsgroups & Oracle (CART) & & & & 0.19 \\
20 Newsgroups & HERCULES~\cite{petnehazi2025herculeshierarchicalembeddingbasedrecursive} & 0.59 & 0.40 & - & - \\
\bottomrule
\end{tabular}
\label{tab:results_inference}
\end{table}

\begin{table}[t!]
\centering
\caption{Evaluation on held-out sets.}
\begin{tabular} {lllrrrr}
\toprule
Dataset & Model (or LLM-S) & NMI & ARI & ACC & F1 \\
\midrule
Wikipedia & \multirow{2}{*}{LogiPart} & 0.23 & 0.10 & 0.36 & 0.18 \\
Bills &  & 0.17 & 0.05 & 0.04 & 0.02 \\
\midrule
Wikipedia & \multirow{2}{*}{LiSa~\cite{liu2025lisa}} & 0.72 & 0.67 & 0.75 & - \\
Bills & & 0.51 & 0.42 & 0.59 & - \\
\midrule
Wikipedia & \multirow{2}{*}{TopicGPT~\cite{topicgpt2024}} & 0.70 & 0.60 & 0.74 & - \\
Bills & & 0.51 & 0.40 & 0.57 & - \\
\bottomrule

\end{tabular}

\label{tab:results_wiki}
\end{table}

\subsection{Path-guided Generation and Inverse Logic Validation}
To assess whether the induced symbolic predicates serve as stable, executable semantic constraints, we implement an \textit{Inverse Logic Validation} protocol. This procedure, inspired by generative consistency checks in TopicGPT~\cite{topicgpt2024} and path-based synthesis in TELEClass~\cite{zhang2025}, tests whether the generated logic is robust enough to guide a separate LLM in synthesizing representative data that can be correctly re-classified by the system.

We compare two generative regimes using a held-out model (\textsc{qwen3:14b}) to avoid self-evaluative bias: (i)  \textbf{Path-driven:} The LLM synthesizes 200 documents conditioned strictly on the sequence of logical predicates leading to a specific leaf node; (ii) \textbf{Few-shot (Exemplar-driven):} The LLM is provided with five random documents from a leaf as prompts to generate 200 similar texts.

Each synthetic document is then routed through the tree using the NLI inference pipeline. We report three metrics in Table~\ref{tab:results_generation}: (i) \textit{Average Correct Node Depth} (CND), the mean depth (relative to the path length) reached before a routing error occurs; (ii) \textit{Path-Fidelity} ($P_\text{path}$), the fraction of documents that successfully return to their intended leaf; and (iii) \textit{Node-Fidelity} ($P_\text{node}$), a normalized measure of per-node decision accuracy. Because reaching a leaf requires a sequence of independent binary decisions, we define Node-Fidelity as $\sqrt[H-1]{P_\text{path}}$, where $H$ is the tree height and $H-1$ represents the number of decision boundaries (edges) in the path.

\begin{table}[t!]
\centering
\caption{Inverse Logic Validation results comparing path-driven and few-shot (exemplar-driven) data synthesis.}
\label{tab:results_generation}

\begin{tabular}{llccc}
\toprule
Dataset & Method & CND & $P_{\text{path}}$ & $P_{\text{node}}$ \\
\midrule
\multirow{2}{*}{AG-News} & Path-Driven & 0.63 & 0.46 & 0.87 \\
& Few-Shot    & 0.18 & 0.04 & 0.52 \\
\midrule
\multirow{2}{*}{20 Newsgroups} & Path-Driven & 0.55 & 0.46 & 0.86 \\
& Few-Shot    & 0.12 & 0.06 & 0.57 \\
\midrule
\multirow{2}{*}{Wikipedia} & Path-Driven & 0.85 & 0.80 & 0.96 \\
& Few-Shot  & 0.15 & 0.05 & 0.55 \\

\midrule
\multirow{2}{*}{Bills} & Path-Driven & 0.80 & 0.68 & 0.93 \\
& Few-Shot  & 0.19 & 0.11 & 0.64  \\

\bottomrule
\end{tabular}
\end{table}

The results demonstrate that path-driven generation provides a significantly higher degree of semantic control compared to few-shot prompting. Notably, the Node-Fidelity for the path-driven regime reaches 0.87, suggesting that the generated predicates function as highly reliable logical gates that maintain their meaning across different models. Conversely, the few-shot regime yields a Node-Fidelity of 0.52, indicating that raw document similarity is no more effective than a stochastic coin flip for navigating deep hierarchical taxonomies. These results demonstrate not merely interpretability, but that predicates function as reusable, model-agnostic decision operators. This confirms that \textsc{LogiPart}'s symbolic predicates capture a distilled logic that generalizes beyond the specific examples found in the discovery set.

\subsection{Predicate quality assessment}
To assess the semantic integrity of the induced predicates beyond thematic alignment metrics, we conduct a qualitative audit using an LLM-as-a-judge protocol. This evaluation probes whether the generated predicates capture non-trivial, intent-level semantic axes rather than replicating surface-level topic boundaries. We employ \textsc{Gemini-2.5-Flash} as an external judge model due to its larger parameter count and strong performance on recent reasoning-oriented benchmarks, and to avoid architectural or training overlap with the locally hosted models used for predicate generation. For each predicate, the judge is presented with a fixed set of seven yes/no questions designed to evaluate redundancy, semantic scope, intent abstraction, clarity, and practical utility. Each question is evaluated independently across five stochastic runs with temperature set to 0.4. Repeated sampling mitigates random hallucinations and instability in single-pass judgments, while controlled stochasticity ensures that predicates with genuinely ambiguous or underspecified semantics yield mixed responses rather than artificially confident scores. For each question, we report the fraction of affirmative (``Yes'') responses across runs, producing a score in $[0,1]$. Reported values are averaged across predicates generated for each dataset and model. Questions q1 and q7 are negatively oriented; all remaining questions are positively oriented, where higher values indicate stronger predicate quality. This protocol does not aim to replace human evaluation, but serves as a scalable proxy for assessing whether the discovered predicates plausibly correspond to stable, interpretable semantic criteria. The prompt includes the list of categories in the dataset and questions shown in Figure~\ref{fig:rubric}. The corresponding results are shown in Table~\ref{tab:results_llm}.

\begin{figure}[h]
\centering
\fcolorbox{blue}{gray!10}{%
  \parbox{0.95\linewidth}{%
\footnotesize
q1. Is this predicate largely redundant with simply assigning documents to one of these categories? In other words, does it mostly replicate at least one of the gold topic boundaries (e.g., it would assign almost the same documents as one of the categories)?

q2. Does this predicate apply meaningfully (i.e., creates a non-trivial split of documents) across more than one of these categories?

q3. Does knowing the answer to this predicate (True/False) provide additional, relevant information within at least one (or more) of these categories that the category label alone does not capture?

q4. Does the predicate denote a human intention underlying the text, rather than a linguistic characteristic like the presence of a specific word, phrase, entity, or pattern?

q5. Is the predicate sufficiently precise and unambiguous that a diverse set of human readers would reliably agree ($\geq 80 \% $ agreement expected) on True/False labels for unseen documents without needing extra definitions or context?

q6. Does this predicate have clear potential for real-world applications in information retrieval, content recommendation, or data organization (e.g., could it meaningfully filter/group news articles for users)?

q7. Does this predicate denote a meaningful subset of exactly one of the topics (i.e., it is not redundant with the topic label, but it only applies within a single topic)?

}}

\caption{Rubric for the LLM-as-a-judge evaluation}
\label{fig:rubric}
\end{figure}
\begin{table*}[t]
\centering
\caption{Results for LLM-as-a-judge evaluation}
\begin{tabular}{llrrrrrrr}
\toprule
Dataset & Model & q1 (redundant) & q2 (cross-topic) & q3 (added value) & q4 (intent) & q5 (clarity) & q6 (utility) & q7 (single-topic) \\
\midrule
agnews & llama3.2:3b & 0.60 & 0.70 & 0.95 & 0.95 & 0.45 & 0.95 & 0.25 \\
agnews & llama3.1:8b & 0.57 & 0.73 & 0.87 & 0.81 & 0.68 & 1.00 & 0.08 \\
agnews & qwen3:8b & 0.27 & 1.00 & 1.00 & 0.84 & 0.64 & 1.00 & 0.00 \\
agnews & qwen3:14b & 0.73 & 0.48 & 0.81 & 0.89 & 0.92 & 1.00 & 0.31 \\
agnews & gpt-oss:20b & 0.63 & 0.56 & 0.99 & 0.79 & 0.85 & 1.00 & 0.39 \\
newsgroups & gpt-oss:20b & 0.37 & 0.99 & 1.00 & 0.97 & 0.85 & 1.00 & 0.00 \\
wikipedia & gpt-oss:20b & 0.17 & 0.95 & 0.97 & 0.65 & 0.61 & 1.00 & 0.01 \\
bills & gpt-oss:20b & 0.16 & 0.96 & 1.00 & 0.79 & 0.77 & 1.00 & 0.04 \\
\bottomrule
\end{tabular}
\label{tab:results_llm}
\end{table*}

To lower API costs, we constrained the evaluation to the first 4 levels of the trees. The results, presented in Table~\ref{tab:results_llm}, provide a quantitative bridge between low thematic alignment scores and high exploratory utility. Across all datasets, the larger models (14B/20B) achieve higher scores in \textit{added value} ($q_3=0.81$/$1.0$) and \textit{practical utility} ($q_6=1.0$), confirming that the system successfully surfaces information not captured by ground-truth labels alone. 

The results for the complex domains of \textbf{Bills} and \textbf{Wikipedia} validate the ``alignment gap'' hypothesis: these datasets show the lowest redundancy with gold topic boundaries ($q_1=0.16$/$0.17$) and the highest \textit{cross-topic applicability} ($q_2=0.95$/$0.96$). This demonstrates that in high-entropy environments, \textsc{LogiPart} intentionally bypasses thematic subject matter to identify functional or ideological axes, such as policy intent, that cross-cut multiple administrative categories. Furthermore, we observe a clear \textit{reasoning threshold} in predicate clarity ($q_5$); while the 2B and 8B models exhibit inconsistent clarity scores ($0.45$--$0.68$), the 14B and 20B models achieve stable, high-agreement predicates ($0.85$--$0.92$). These findings confirm that \textsc{LogiPart} effectively distills the opaque signal of the embedding manifold into human-meaningful, intent-level taxonomic criteria that generalize beyond simple keyword matching.

\subsection{Usage and Scalability Analysis}
\label{sec:usage}

\begin{figure}[h!]
    \centering
    \begin{subfigure}[t]{0.49\linewidth}
        \centering
        \includegraphics[width=\linewidth]{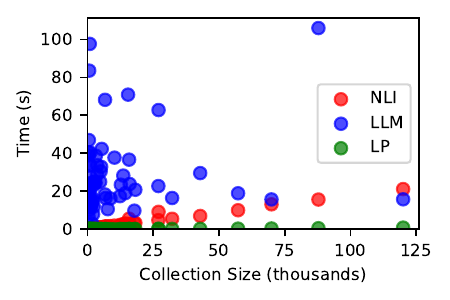}
        \caption{Wall-clock time (seconds) spent on each component: LLM hypothesis generation, NLI evaluation, and label propagation (LP).}
        \label{fig:wallclock}
    \end{subfigure}
    \hfill
    \begin{subfigure}[t]{0.49\linewidth}
        \centering
        \includegraphics[width=\linewidth]{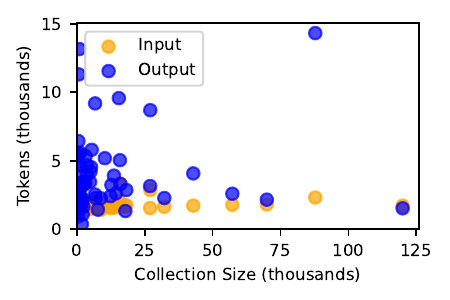}
        \caption{Number of input and output tokens consumed by LLM hypothesis generation per node.}
        \label{fig:n_tokens}
    \end{subfigure}
    \caption{Overall results comparing time spent and token usage per node.}
    \label{fig:results}
\end{figure}

To illustrate the practical scalability of the proposed framework, we analyze resource consumption during hierarchy construction using our preferred configuration (gpt-oss-20B with Vote-K sampling for hypothesis generation and K-Means for NLI assignment) on the AG-News discovery set. We measured wall-clock time per node (Figure~\ref{fig:wallclock}) and LLM token usage (input and output) per node (Figure~\ref{fig:n_tokens}) across nodes of varying collection sizes.

As shown in Figure~\ref{fig:wallclock}, time spent on NLI evaluation increases linearly with subtree size, as expected since NLI is applied to a sampled subset followed by propagation. Critically, label propagation is orders of magnitude faster than full NLI, confirming that this hybrid discriminative approach effectively scales hypothesis evaluation to large collections.

LLM hypothesis generation dominates overall runtime but remains largely independent of collection size. This desirable behavior arises because the input prompt is constructed from a fixed-size sample, keeping input token count approximately constant across nodes (Figure~\ref{fig:n_tokens}). Output token count, however, varies substantially due to differences in the model's reasoning chain length when formulating hypotheses.

LLM runtime correlates almost perfectly with output tokens (Pearson $ r = 0.99, p < 10^{-80}, n>50$), indicating variability stems from reasoning length rather than input scale. This underscores that hypothesis quality benefits from the LLM's capacity for extended chain-of-thought, while the overall token complexity per level remains \( O(1) \) with respect to corpus size — a key advantage over recursive full-corpus LLM conditioning approaches that scale linearly.

These measurements validate the design premise: by confining expensive generative LLM calls to compact, representative samples and delegating bulk assignment to fast discriminative models, the framework achieves scalable, interpretable hierarchy construction without prohibitive computational cost.

\section{Discussion and Limitations}
\label{sec:discussion}

\textbf{Interpretability vs. Geometric Optimality:} Our results highlight a fundamental tension between geometric optimality and semantic interpretability. \textsc{LogiPart} explicitly trades off thematic ``recovery'' (high NMI/F1) for taxonomic discovery, that is, identifying functional or intent-level axes (e.g., ``Public Welfare vs. Individual Benefit'') that cross-cut administrative categories. The high node-fidelity (up to 0.96) and judge utility and added value scores confirm that this ``alignment gap'' is not a failure of clustering, but a successful surfacing of latent dimensions invisible to traditional metrics.

\textbf{Economic Scalability and Steerability:} A critical differentiator of our framework is its scaling law. SOTA generative frameworks like \textsc{TopicGPT}~\cite{topicgpt2024} incur a linear $O(N)$ generative cost; at current API pricing, processing a 14,000-document corpus is estimated to cost over \$100 per run. By achieving constant $O(1)$ generative complexity per node relative to $N$, \textsc{LogiPart} reduces the marginal cost of scaling to zero. Even if $O(N)$ frameworks are executed locally, that would require several seconds per document for generative inference, which quickly scales to week- or month-long execution times. Furthermore, this decoupled architecture enables \textit{interactive steerability}: because generative calls are confined to small samples, users can manually edit or refine a specific logical predicate and re-propagate labels across the full corpus. This iterative refinement is financially and computationally prohibitive in frameworks that require per-document LLM conditioning.

\textbf{Dependence on NLI and Text Structure:} The reliability of the NLI model remains a central dependency. While NLI is significantly more efficient than generative LLMs, it is susceptible to miscalibration on abstract hypotheses involving discourse function. This risk is partially mitigated by our recursive acceptance rules, which reject polarized or non-discriminative hypotheses. Additionally, our 350-word sampling assumes that document semantics are front-loaded. While this holds for news and encyclopedic texts~\cite {marchionini2006}, it may fail for long-form narratives or scientific papers, where critical distinctions emerge later. Handling such domains may require future integration with adaptive chunk selection or hierarchical summarization.

\textbf{Positioning Relative to Decision Trees:} \textsc{LogiPart} can be viewed as a text-native analogue of decision tree induction, where splits are defined by natural-language predicates rather than feature thresholds. Unlike classical decision trees, however, the predicates are not optimized for information gain but are proposed by an LLM and validated post hoc. This paradigm prioritizes human navigability and semantic abstraction over strict statistical optimality, aligning the method more closely with exploratory sensemaking tools than with predictive classifiers.

\section{Conclusion}
\label{sec:conclusion}

We presented \textsc{LogiPart}, a scalable, hypothesis-first framework for interpretable hierarchical text discovery. By decoupling taxonomic discovery from full-corpus labeling, \textsc{LogiPart} achieves a constant $O(1)$ generative token complexity per node related to the corpus size, shifting the computational burden to efficient NLI-based propagation. This design enables frontier-level exploratory analysis on consumer-grade hardware with zero marginal API cost. Beyond the specific architecture, we contribute a multi-stage validation framework for exploratory discovery, combining Inverse Logic Validation to ensure structural stability and a functional meta-evaluation rubric to measure semantic utility.

While traditional thematic metrics reveal an ``alignment gap'' in complex corpora like US Bills, our qualitative audit and inverse-logic validation confirm that this reflects the successful discovery of orthogonal, functional dimensions (e.g., policy intent) rather than a failure of clustering. With an average per-node routing accuracy of up to 96\%, \textsc{LogiPart} demonstrates that symbolic logic provides a more stable and navigable foundation for exploration than opaque embedding geometry. Furthermore, by treating the LLM and NLI components as modular generative and discriminative backbones, \textsc{LogiPart} offers a model-agnostic architecture that can seamlessly integrate future advancements in transformer-based neural architectures.

Beyond performance, the steerable nature of the \textsc{LogiPart} architecture enables rapid, iterative taxonomic refinement, empowering researchers to interactively map large-scale corpora within realistic computational constraints. Taken together, our results suggest that the future of large-scale sensemaking lies in combining the generative intuition of local LLMs with the discriminative efficiency of symbolic inference.

\section*{Acknowledgements}
ChatGPT, Gemini, and Grok were used to improve the writing style throughout the text.

\bibliographystyle{ieeetr}
\bibliography{references}

\end{document}